\documentclass[preprint,12pt]{elsarticle}

\usepackage{amssymb}
\usepackage{algorithm}
\usepackage{algpseudocode}
\usepackage{makecell} 
\usepackage{url}
\usepackage{rotating}
\usepackage{tabularx} 
\usepackage{booktabs}
\usepackage{makecell}
\usepackage{multirow}
\usepackage{amsmath}
\usepackage{amsthm}
\usepackage{subcaption}
\usepackage{bigstrut,multirow,rotating,booktabs}
\newtheorem{definition}{Definition}

\begin{document}

\begin{frontmatter}



\title{LSPI: Heterogeneous Graph Neural Network Classification Aggregation Algorithm Based on Size Neighbor Path Identification}

\author{Yufei Zhao}
\ead{yfzhao@sdust.edu.cn}

\author{Shiduo Wang}
\ead{wangshiduo@sdust.edu.cn}

\author{Hua Duan\corref{cor1}}
\ead{huaduan59@163.com}

\cortext[cor1]{Corresponding author. Email address: huaduan59@163.com}

\affiliation{organization={College of Mathematics and Systems Science, Shandong University of Science and Technology},
	city={Qingdao},
	postcode={266590},
	state={Shandong},
	country={China}}

\begin{abstract}
Existing heterogeneous graph neural network algorithms (HGNNs) mostly rely on meta-paths to capture the rich semantic information contained in heterogeneous graphs (also known as heterogeneous information networks (HINs)), but most of these HGNNs focus on different ways of feature aggregation and ignore the properties of the meta-paths themselves. This paper studies meta-paths in three commonly used data sets and finds that there are huge differences in the number of neighbors connected by different meta-paths. At the same time, the noise information contained in large neighbor paths will have an adverse impact on model performance. Therefore, this paper proposes a Heterogeneous Graph Neural Network Classification and Aggregation Algorithm Based on Large and Small Neighbor Path Identification(LSPI). LSPI firstly divides the meta-paths into large and small neighbor paths through the path discriminator , and in order to reduce the noise interference problem in large neighbor paths, LSPI selects neighbor nodes with higher similarity from both topology and feature perspectives, and passes small neighbor paths and filtered large neighbor paths through different graph convolution components. Aggregation is performed to obtain feature information under different subgraphs, and then LSPI uses subgraph-level attention to fuse the feature information under different subgraphs to generate the final node embedding. Finally this paper verifies the superiority of the method through extensive experiments and also gives suggestions on the number of nodes to be retained in large neighbor paths through experiments. The complete reproducible code adn data has been published at:  \url{https://github.com/liuhua811/LSPIA}.
\end{abstract}



\begin{keyword}
	 Heterogeneous graph neural network \sep Node filtering \sep Graph embedding \sep Graph representation learning
\end{keyword}

\end{frontmatter}


\section{Introduction}

With the rapid development of neural networks, the application of neural networks in the real world is rapidly gaining popularity. However, traditional neural networks only work on Euclidean spatial data, but non-Euclidean spatial data are also prevalent in the real world, such as heterogeneous graph. In order to address the feature capture ability of neural networks for non-Euclidean spatial data such as heterogeneous graph, heterogeneous graph neural networks have attracted the attention of a wide range of researchers in recent years. Currently, heterogeneous graph neural networks have been applied in areas such as academic networks \cite{bib1,bib2}, transportation systems \cite{bib3}, drug response \cite{bib4} and physical systems \cite{bib5}.Therefore, doing a good job in feature mining of heterogeneous graph neural networks has important application value and economic significance.

Many existing heterogeneous graph neural networks have achieved excellent performance on real-world heterogeneous graphs \cite{bib6,bib7,bib8,bib9}, due to the heterogeneity of heterogeneous graphs, the same type of nodes tend not to be directly connected, so these models mostly capture the same type of neighbor nodes with the help of meta-paths. Meta-path is a unique form of connectivity in HIN, through which the same type of neighbors with different semantic connectivity relationships can be captured.

\begin{figure}[!h]
	\centering
	\includegraphics[width=\textwidth]{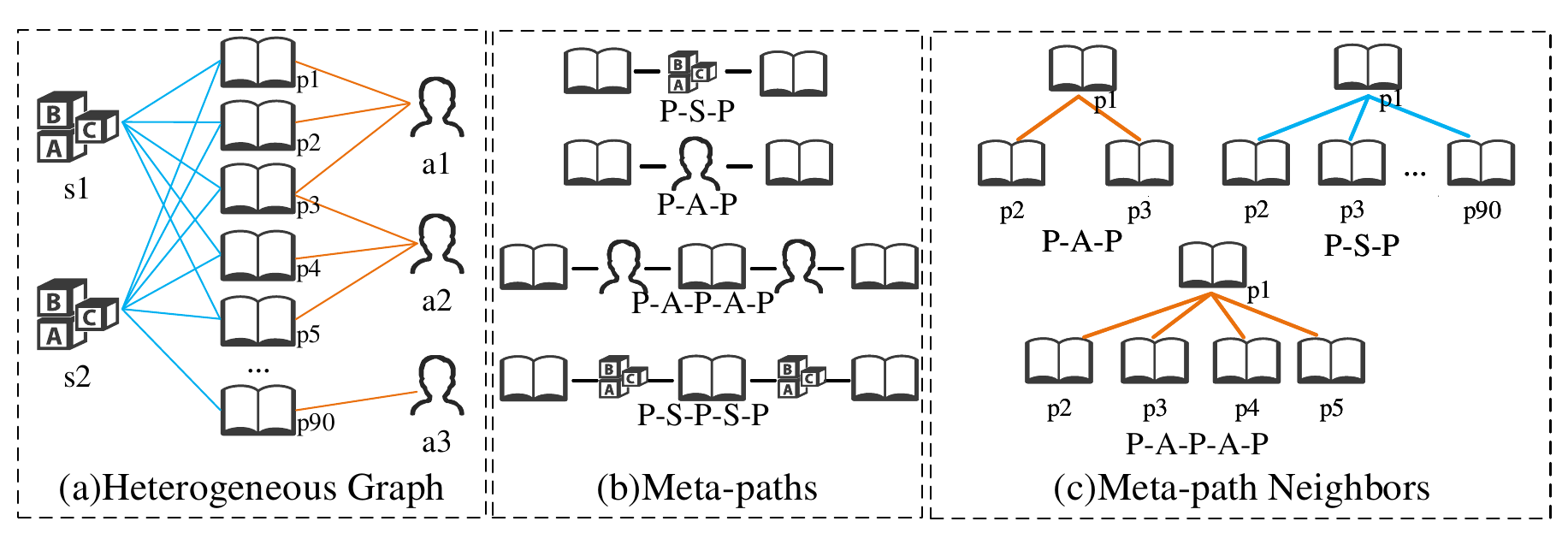}
	\caption{Example of a heterogeneous graph (ACM). (a) A heterogeneous graph ACM composed of three node types, where P denotes paper, A denotes author, and S denotes subject; (b) Several meta-paths with different semantics and different lengths in the heterogeneous graph ACM; (c) Neighbors based on the three meta-paths; for simplicity, the following content only expresses meta-paths in the order of node connections, such as P-A-P abbreviated as PAP.}
	\label{fig1}
\end{figure}

\begin{figure}[!h]
	\centering
	\includegraphics[width=\textwidth]{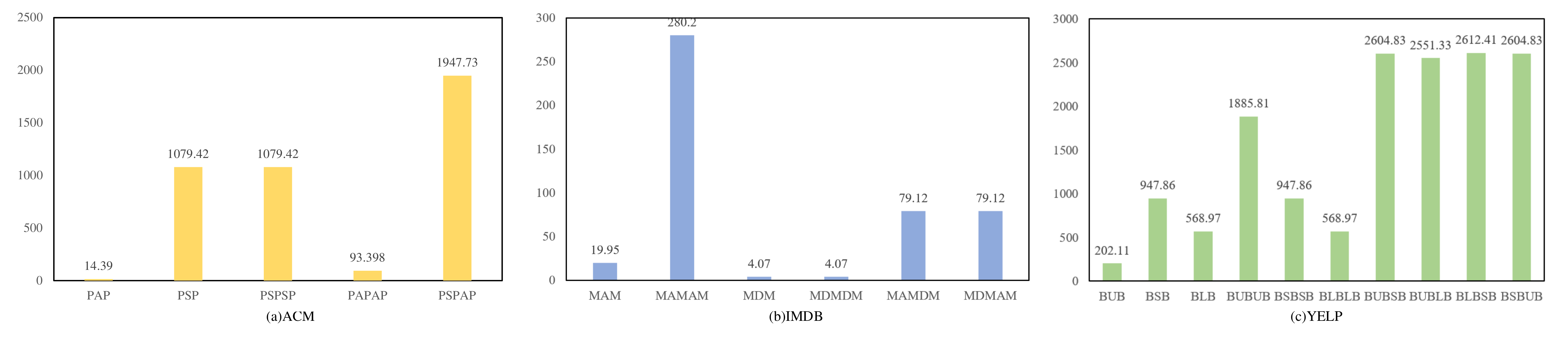}
	\caption{Mean difference in the number of node neighbors under different meta-paths.}
	\label{fig2}
\end{figure}

In this paper, we find that there is a huge difference in the number of neighbors of meta-paths with different semantics or different lengths. Taking the two meta-paths with the same length (PAP and PSP) in Figure\ref{fig1} as an example, the number of neighbors of the papers connected through the same authors shows a huge difference from the number of neighbors of the papers connected through the same topics; and for two semantically similar meta-paths (PAP and PAPAP) with different lengths, the number of connected nodes often increases exponentially as the length increases. Based on this analysis, this paper further studies the difference in the average number of node neighbors under different meta-paths in three commonly used data sets (ACM, IMDB, and Yelp), and the results are shown in Fig.2. It can be found from Figure\ref{fig2} that in the ACM data set, the difference in the average number of neighbors between PAP and PSP, meta-paths of the same length, reaches 75 times, and the difference in the average number of neighbors between PAP and PAPAP, meta-paths with similar semantics but different lengths, also reaches 52 times. In IMDB, the difference in the number of neighbors between meta-paths of the same length(MAM and MDM) is about 5 times, but the difference in the number of neighbors between semantically similar meta-paths with different lengths is also 14 times (MAM and MAMAM). The number of meta-path neighbors with different semantics and different lengths in Yelp also shows huge differences. For convenience, this article refers to a meta-path with a large number of neighboring nodes as a Large Neighbor Path (abbreviated as LargePath), and a meta-path with only a few neighboring nodes as a Small Neighbor Path (abbreviated as SmallPath). 

\begin{figure}[!h]
	\centering
	\includegraphics[width=\textwidth]{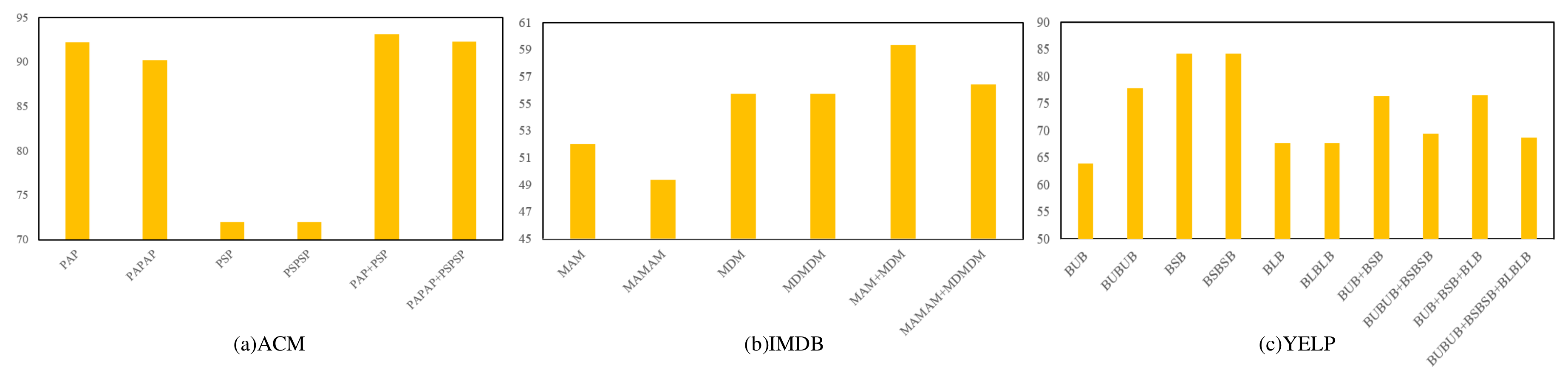}
	\caption{Accuracy of HAN with different meta-paths.}
	\label{fig3}
\end{figure}

It is unscientific to aggregate meta-paths with huge differences in the number of neighbors in the same way without distinction. Especially for LargePaths, there must be noise information in a large number of neighbor nodes. This paper verifies this conclusion through the advanced SOTA model HAN, and the results are shown in Figure\ref{fig3}. Since different semantics have different importance, this article only compares the performance between meta-paths with similar semantics but different lengths. It can be found that on the ACM data set, as the number of neighbors increases, the accuracy of meta-path PAPAP decreases by up to 2\% compared with PAP, and the accuracy of meta-path PAPAP+PSPSP also decreases by about 2\% compared with PAP+PSP. This problem appears again on IMDB (MAMAM and MAM, MAMAM+MDMDM and MAM+MDM). Although the accuracy of BUBUB on Yelp has increased compared with BUB, this is due to the fact that the number of BUB neighbor nodes is too small and the complete neighbor information cannot be captured. After the number of neighbors increases, the accuracy rate under the large neighbor path combination drops significantly (BUBUB+BSBSB and BUB+BSB, BUBUB+BSBSB+BLBLB and BUB+BSB+BLB).

However, to the best of the our knowledge, existing HGNNs mostly focus on different aggregation algorithms based on meta-paths, and there has not yet been any dedicated work to address the huge difference in the number of neighbors among different meta-paths. Although RoHe proposes an attention purification mechanism against adversarial attacks, it focuses on the network purification problem when facing adversarial attacks rather than the neighbor differences between different meta-paths. Based on the above analysis this paper proposes a Heterogeneous Graph Neural Network Classification and Aggregation Algorithm Based on Large and Small Neighbor Path Identification (LSPI). LSPI is divided into three parts: path discriminator, intra-path aggregation and subgraph-level attention aggregation. Specifically, after inputting Heterogeneous Graph into LSPI, the model firstly divides the graph topology into large neighbor paths(LargePaths) and small neighbor paths(SmallPaths) through the path discriminator, and for the LargePaths, LSPI uses topological priors and node feature similarities to select and aggregate neighbor nodes with the highest topological probability and feature similarity in the neighbor set. For SmallPaths LSPI uses subgraph aggregation to aggregate meta-path subgraphs to obtain feature embeddings under specific subgraphs. Finally, LSPI fuses the obtained LargePaths embedding and SmallPaths embedding through subgraph-level attention to generate the final node representation.

Specifically, the contributions of this paper are as follows:

\begin{itemize}
	\item 	The paper is the first to address the problem of huge differences in the number of neighbors of meta-paths in heterogeneous graph neural networks, and analyzes the impact of noise information in large neighbor paths on the performance of the model. 
	\item	The paper proposes a Heterogeneous Graph Neural Network Classification and Aggregation Algorithm Based on Large and Small Neighbor Path Identification, using a path discriminator to divide the meta-paths into large neighbor paths and small neighbor paths, and feature aggregation by different paths.
	\item 	In large neighbor path aggregation, LSPI selects the neighbor node with the highest topological relationship and feature similarity from topological probability and node feature similarity.
	\item   The superiority of LSPI on various tasks is verified by different experiments on three real-world datasets, while this paper explores how many neighbor nodes are retained under the large neighbor paths that are most conducive to improving the performance of the model.
\end{itemize}

\section{Related Work}
In this subsection we summarize some existing related work, including heterogeneous graph neural networks and large neighbor path node selection.

\textbf{Heterogeneous Graph Neural Networks.} HAN\cite{bib10} is a pioneering work on heterogeneous graph neural networks, which uses manually designed meta-paths and the idea of hierarchical aggregation to capture semantic information within and between meta-paths; however, considering that HAN ignores the information of intermediate nodes when aggregating within meta-paths, MAGNN\cite{bib11} further uses relational rotary encoders to aggregate meta-path instances in order to avoid the intermediate node loss; HPN\cite{bib12} proposed a novel heterogeneous graph propagation network to capture higher-order semantics in order to alleviate the degradation phenomenon in deep HGNNs, so that it can appropriately absorb local semantics during semantic propagation to avoid the semantic confusion problem. HGT\cite{bib13} designed relevant parameters for node and edge types to characterize the heterogeneous attention on each edge. This allows HGT to maintain dedicated representations for different types of nodes and edges, while also introducing temporal encoding techniques to capture the dynamic changes in the graph. HetGNN\cite{bib14} uses a heterogeneous neighbor sampling strategy for nodes with the same attributes and different types in a heterogeneous graph using two aggregation methods to capture the structural information of the heterogeneous graph and the content information of each node; HetSANN\cite{bib15} leverages the structural information of heterogeneous graphs to enhance node representation learning by employing a structure-aware approach to handle interactions between different types of nodes. GCNH\cite{bib16} uses a learnable importance coefficient to balance the contributions of central nodes and neighboring nodes, obtaining independent representations for the combination of a node and its neighbors. BPHGNN\cite{bib17} proposes a depth and breadth behavior pattern aggregation method, which automatically captures local and global relevant information, adaptively learning the importance of various behavior patterns for multi-layer heterogeneous network representation learning. SR-HGN\cite{bib18} captures feature information from both relational and semantic aspects, and generates feature embeddings that fuse relational and semantic aspects.

However, the aforementioned models primarily focus on different feature aggregation strategies and do not propose effective measures to address the noise problem in the large number of neighboring nodes, which can result in suboptimal outcomes.

\textbf{Large Neighbor Path Node Selection.} Due to the complexity of the graph structure, how to select a more valuable subset of nodes for aggregation from the noise-filled large neighbor paths is a challenging task. To the best of the authors' knowledge, there has been no dedicated research specifically addressing this issue, but many works have made valuable attempts at neighbor selection. GCN\cite{bib19} as a seminal work in the field of graph neural networks, treats all the first-order neighbors as direct aggregation objects and extends the aggregation to higher orders by superimposing layers; GAT\cite{bib20} uses an attention mechanism to dynamically select the neighbors of a node and assigns different weights based on the degree of interactions between the nodes to better capture the associative relationships between the nodes in the graph structure; GraphSAGE\cite{bib21} uses random walks to generate node sequences from neighboring nodes for aggregation, avoiding the aggregation of all neighboring nodes to reduce the impact of noisy information. However, this method is inherently random; HetGNN\cite{bib14} uses a heterogeneous neighbor sampling strategy based on restarting random walk (RWR) to collect all types of neighbors for each vertex, and then aggregates the information of different types of neighbor nodes in order to learn better node representations; RoHe\cite{bib22} proposes a novel approach that employs an attention purification mechanism to shield against malicious neighbor nodes during adversarial attacks. However, it primarily targets malicious nodes in adversarial scenarios rather than addressing noisy information in large neighborhood paths. DCNN\cite{bib23} employs the concept of diffusion kernels to obtain feature representations for each node based on the diffusion process to determine neighbor nodes; AGCN\cite{bib24} performs neighborhood sampling of nodes at different scales through multi-scale neighborhood sampling; HetSANN\cite{bib15} considers the structural relationships between nodes when selecting neighbor nodes and dynamically adjusts the selection strategy based on node types and edge types to ensure that the chosen neighbor nodes best reflect the structural relationships between nodes. While these works select neighbor sets from different perspectives, they do not address the noise problem in large neighborhood paths. Therefore, their selection strategies will be adversely affected when facing large neighbor paths with noise interference.

\section{Preliminaries}

\begin{definition}[Heterogeneous Graph \cite{bib10}]
A HIN can be represented as ${\mathcal{G}=\{\mathcal{V},\mathcal{E},\mathcal{A},\mathcal{R}}\}$, consisting of the set of objects ${\mathcal{V}}$ and the set of edges ${\mathcal{E}}$ as well as the set of object types ${\mathcal{A}}$ and the set of edge types combined with ${\mathcal{R}}$, where ${\mathcal{A}=\{\mathcal{A}_i|i\geq1\}, \mathcal{R}=\{\mathcal{R}_i|i\geq1\}}$, and ${|\mathcal{A}|+|\mathcal{R}|>2}$.

\end{definition}

\begin{definition}[Meta-paths \cite{bib10}]\label{de2}
A meta-path ${\Phi}$  defined as a path connected by different objects and relations in the form of \(A_1 \xrightarrow{R_1} A_2 \xrightarrow{R_2}  \cdots \xrightarrow{R_l}  A_{l+1}\)  . A meta-path describes a composite relationship \(R = R_1 \circ R_2 \circ \cdots \circ R_l\)  between node types ${A_1}$  and ${A_{l+1}}$ .
	
\end{definition}

\begin{definition}[Meta-path-based Neighbors \cite{bib11}]\label{de3}
For a given node v and meta-path ${\Phi}$ in a heterogeneous graph, a meta-path neighbor ${N_v^\mathrm{\Phi}}$ is defined as the set of nodes connected to node v via a meta-path ${\Phi}$.
	
\end{definition}

\begin{definition}[Meta-path-based Neighbors \cite{bib22}]\label{de4}
The transit probability based on meta-path is defined as the probability from node v to node u along the meta-path ${\Phi}$, and the transit probability is a manifestation of connectivity within the meta-path.
	
\end{definition}

\section{Model}

This section will introduce a new LSPI model. LSPI takes a given HIN ${\mathcal{G}}$ and the attribute matrix ${X_{A_i}\in\mathbb{R}^{|\mathcal{V}_{A_i}|\times d_{A_i}}}$ of node type ${A_i\in\mathcal{A}}$  as input, learning a mapping function ${f:\mathcal{V}\rightarrow\mathbb{R}^d}$ where ${d\ll|\mathcal{V}|}$, to capture the rich structural and semantic information involved in ${\mathcal{G}}$, and generate the final feature representations for downstream tasks.

\begin{figure}[!h]
	\centering
	\includegraphics[width=\textwidth]{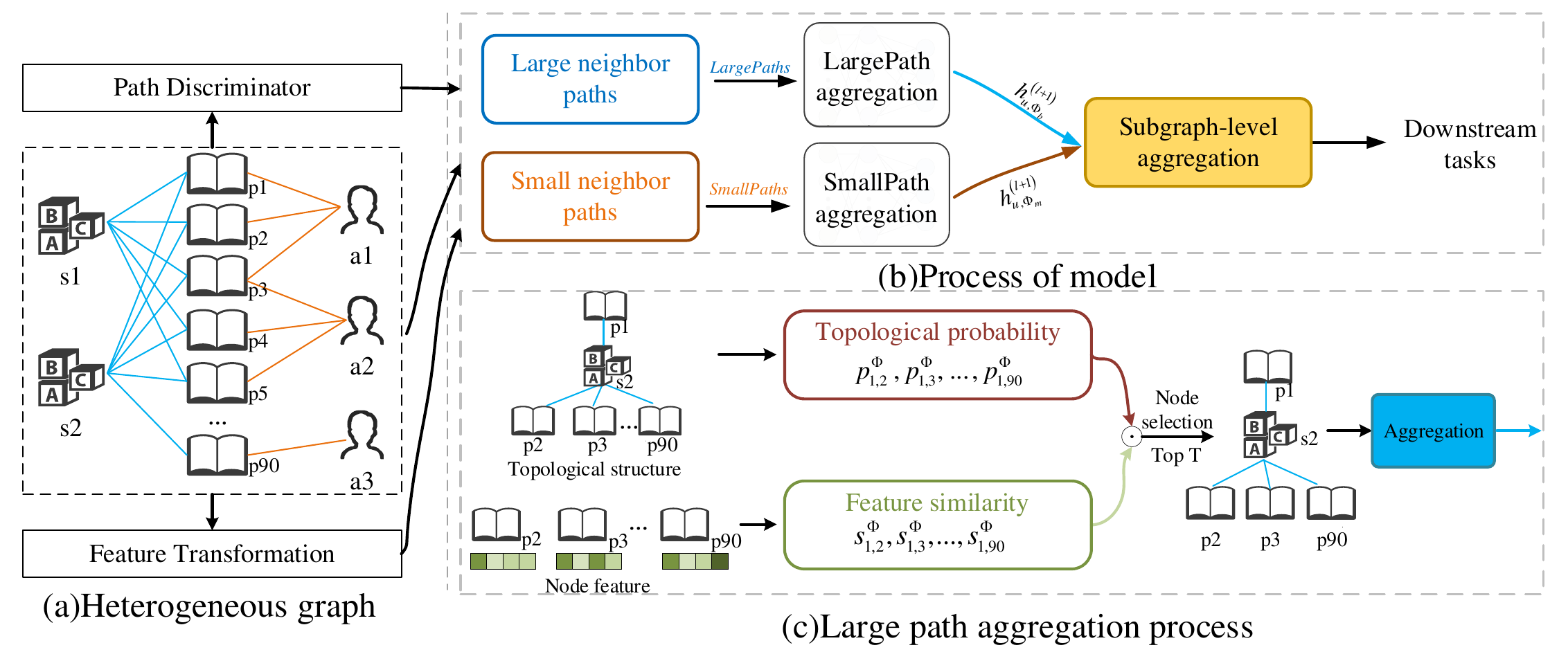}
	\caption{LSPI framework structure.}
	\label{fig4}
\end{figure}

Specifically,as shown in Figure\ref{fig4}, LSPI first uses a path discriminator to divide meta-paths according to the topology of the graph. The path discriminator divides meta-paths into LargePaths and SmallPaths by calculating the percentage change in degree values between different paths. For LargePaths, LSPI finds out the highest correlation nodes from many neighbor nodes from both topology and feature perspectives for aggregation in order to shield noise harassment. From a topological perspective, LSPI uses transition probability priors to calculate the transit probability of nodes. From a feature perspective, LSPI calculates the feature similarities of all nodes. Then LSPI selects the node with the highest transit probability and feature similarity from the large neighbor path for aggregation. For SmallPaths, LSPI uses convolution operation to capture feature information of specific subgraphs. Finally, LSPI employs graph-level attention to aggregate the features of all LargePaths with SmallPaths to obtain the final node embedding.

\subsection{Path Discriminator}

Path Discriminator first calculates the total degree value of all nodes of each meta-path as shown in Equation 1:
\begin{equation}\label{eq1}
	D_{\Phi_m}=\sum_{i=1}^{\left|\mathcal{V}_A\right|}\sum_{j=1}^{\left|\mathcal{V}_A\right|}d_{i,j}
\end{equation}

In the above equation,${D_{\Phi_m}}$ is the degree sum of all the nodes under the meta-path ${\Phi_m}$, ${d_{i,j}}$ denotes the value of the ${j}$-th element of the ${i}$-th row of ${\Phi_m}$, and ${\left|\mathcal{V}_A\right|}$ denotes the number of nodes of the target type.

After the above calculation, the degree value set ${D_\Phi=\left\{D_{\Phi_i}|i\in\left(1,...,p\right)\right\}}$ of all meta-paths is obtained, where p is the number of meta-paths. Path Discriminator performs path delineation by calculating the relative differences between the degree values:

\begin{equation}\label{eq2}
D_{\Phi_{min}}=min(D_\Phi)
\end{equation}

\begin{equation}\label{eq3}
	R_{\Phi_i}=\frac{D_{\Phi_i}-D_{\Phi_{min}}}{D_{\Phi_{min}}}\times100%
\end{equation}

After obtaining the relative difference percentage between meta-paths ${R_\Phi=\left\{R_{\Phi_i}|i\in\left(1,...,p\right)\right\}}$, the Path Discriminator divides the meta-paths into large neighbor paths and small neighbor paths according to the relative difference values.

\begin{equation}\label{eq4}
LargePaths=\left\{\Phi_b\in\Phi|R_{\Phi_b}\geq\tau\right\}
\end{equation}

\begin{equation}\label{eq5}
	SmallPaths=\left\{\Phi_s\in\Phi|R_{\Phi_s}<\tau\right\}
\end{equation}

Where ${\tau}$ is the set hyperparameter value, when ${\tau}$ takes 100\% it means that the path degree value in the LargePaths set is at least twice the minimum degree value. In the experimental part, this paper will analyze the value of ${\tau}$.

\subsection{Large Neighbor Path Neighbor Node Selection}

Due to the existence of a large number of neighbor nodes in the large neighbor path, as analyzed in the previous section, it is difficult to truly avoid noise interference when using attention aggregation. Convolution operations treat all nodes as equally important, which inevitably reduces the model's accuracy in the presence of noise. Therefore, filtering out noisy nodes from multiple neighbors to improve the effectiveness of aggregation is an effective method to optimize the aggregation of large neighborhood paths. LSPI selects the neighbors with the highest topological relationships and feature similarity from both topological and feature perspectives for aggregation, in order to avoid noise interference.

At the topological level this paper first calculates the transit probability under meta-path  , for relation ${R_i}$ in ${\Phi}$ there is a transit probability ${P^{R_i}=\left(D^{R_i}\right)^{-1}{Adj}^{R_i}}$, where ${{Adj}^{R_i}}$ and ${D^{R_i}}$ denote the corresponding adjacency matrix and degree of the relation ${R_i}$ respectively matrix. It can be seen that ${P^{R_i}}$ is affected by the number of neighbors, so ${P_{vu}^{R_i}}$ can be understood as the probability of passing through the relation ${R_i}$ from node v to node u. For the meta-path ${\Phi}$ transit probability can be calculated as:

\begin{equation}\label{eq6}
P^\Phi=P^{R_1}P^{R_2}\cdots P^{R_l}
\end{equation}

At the feature level, based on the assumption that nodes with similar features are more important, this paper uses the calculation of feature similarity between node ${v}$ and node ${u}$ as the basis for feature-based judgment. Given the meta-path ${\Phi}$, for the target type node ${V_i\in\mathcal{V}_A}$ has:

\begin{equation}\label{eq7}
h_i^{\prime}=\frac{h_i}{\|h_i\|}
\end{equation}

\begin{equation}\label{eq8}
s_{vu}^\Phi=h_v^{\prime}\cdot h_u^{\prime}+\epsilon 
\end{equation}

where ${s_{vu}^\Phi}$ denotes the feature vector similarity between node v and node u after normalization under a given meta-path ${\Phi}$, ${h_i}$ is the feature vector of node ${V_i}$, and ${\left\|\cdot\right\|}$ is the ${L_2}$ paradigm function, ${\epsilon}$ is a very small value, to avoid the adverse effect on the subsequent operations when the similarity is 0.

LSPI selects doubly better neighbor nodes in large neighbor paths by feature similarity and topological relationship.

\begin{equation}\label{eq9}
	t_{vu}^\Phi=P_{vu}^\Phi\cdot s_{vu}^\Phi 
\end{equation}
${t_{vu}^\Phi\in t_v^\Phi}$ is the topological and feature importance score of the neighbor ${u\in\mathcal{N}_v^\Phi}$ of the node ${v}$. LSPI reconstructs the connectivity under the meta-paths based on the calculated importance score.

\begin{equation}\label{eq10}
C_v^\Phi=Select\_Top(t_v^\Phi,T) 
\end{equation}

${C_v^\Phi}$ is the set of neighbors of node ${v}$ under meta-path ${\Phi}$ with its neighbors ${u\in\mathcal{N}_v^\Phi}$ after similarity selection, ${Select\_Top(\cdot)}$ function can select the top ${T}$ nodes with the highest value in the similarity vector ${t_v^\Phi}$, ${T}$ is a set hyperparameter indicating the number of neighbor nodes retained under the meta-path ${\Phi}$.

\subsection{Intra-path Aggregation}
\subsubsection{Node Feature Conversion}

Considering that different nodes may be located in different feature spaces, we first project different types of nodes to the same dimension for ease of operation. For ${A\in\mathcal{A}}$ type node ${u\in\mathcal{V}_A}$, there are:

\begin{equation}\label{eq11}
	x_u^\prime=W_A \cdot x_u^A
\end{equation}

where ${x_u^A\in\mathbb{R}^{d_A}}$ and ${x_u^\prime\in\mathbb{R}^{d^\prime}}$ are the original features of node u and the projected features after feature conversion respectively. ${W_A\in\mathbb{R}^{d^\prime\times d_A}}$ is the projected features for node type ${A}$ feature projection transformation matrix.

\subsubsection{Large Neighbor Path Aggregation}

After Section 4.2, LSPI then obtains the matrix of connectivity relationships under different meta-paths ${\mathcal{C}=\left(C^{{\Phi}_1},C^{{\Phi}_2},...C^{{\Phi}_p}\right)}$, and for the connectivity relations under different large neighbor paths ${{\Phi}_b\in LargePaths}$ LSPI performs the convolution operation on the subgraph to capture the similar node features after selection.

\begin{equation}\label{eq12}
h_{u,\mathrm{\Phi}_b}^{\left(l+1\right)}=D_{{\Phi}_b}^{-\frac{1}{2}}\widehat{A_{{\Phi}_b}}D_{{\Phi}_b}^{-\frac{1}{2}}h_{u,\mathrm{\Phi}_b}^{\left(l\right)}W_b^{\left(l\right)}
\end{equation}

where ${h_{u,\mathrm{\Phi}_m}^{\left(l+1\right)}}$ is the feature representation of node u at the ${l+1}$ layer of convolution, ${\widehat{A_{{\Phi}_m}}\in\mathbb{R}^{d^\prime\times d^\prime}}$ is the adjacency matrix of ${{\Phi}_m}$ after normalized adjacency matrix, ${D_{{\Phi}_m}}$ is the corresponding degree matrix, ${W\in\mathbb{R}^{d^\prime\times d^\prime}}$ is the learnable parameter matrix, and ${h_{u,\mathrm{\Phi}_m}^{\left(l\right)}}$ is the feature of the node u at the convolution of the ${l}$-th layer representation, where ${h_{u,\mathrm{\Phi}_m}^{\left(0\right)}}$ is the initial node feature of node ${x_u^\prime}$.

\subsubsection{Small Neighbor Path Aggregation}

For small neighbor paths, this article constructs a subgraph based on the small neighbor paths and performs subgraph-level convolution aggregation on the subgraph. For the connection relationship under the small neighbor path ${{\Phi}_m\in SmallPaths}$, LSPI performs a convolution operation on the subgraph to capture the node embedding under the small neighbor path.

\begin{equation}\label{eq13}
	h_{u,\mathrm{\Phi}_m}^{\left(l+1\right)}=D_{\mathrm{\Phi}_m}^{-\frac{1}{2}}\widehat{A_{\mathrm{\Phi}_m}}D_{\mathrm{\Phi}_m}^{-\frac{1}{2}}h_{u,\mathrm{\Phi}_m}^{\left(l\right)}W_m^{\left(l\right)}
\end{equation}

\subsection{Subgraph-level Attention Aggregation}
Since each meta-path contains different semantic information and specific meta-paths can only respond to node information from a single viewpoint, in order to learn more comprehensive node embeddings it is necessary to fuse different meta-paths to enrich the node feature representation. Taking the node embeddings ${H={H^{\mathrm{\Phi}_1},...,H^{\mathrm{\Phi}_p}}}$ as inputs, and LSPI employs subgraph-level attention to aggregate the embeddings learned from different meta-paths.

\begin{equation}\label{eq14}
w_i=\frac{1}{|\mathcal{V}|}\sum_{i\in\mathcal{V}} q_H^T\cdot tanh\left({W_H} \cdot H_i^\mathrm{\Phi}+b_H\right)
\end{equation}

\begin{equation}\label{eq15}
\beta_i=\frac{exp\left(w_i\right)}{\sum_{i=1}^{P}exp\left(w_i\right)}
\end{equation}

\begin{equation}\label{eq16}
	Z=\sum_{i=1}^{P}\beta_iH_i
\end{equation}

Where ${W_H}$  is the weight matrix, ${b_H}$ is the bias vector, ${q_z\in\mathbb{R}^{{1\times d}^\prime}}$ is the graph level attention vector, ${\beta_i}$ is the meta-paths ${\mathrm{\Phi}_i}$ contribution to a particular task. ${Z}$ is the final feature embedding after fusing different semantics.

For semi-supervised node classification task this paper uses cross entropy loss to optimize the model:
\begin{equation}\label{eq17}
	\mathcal{L}=-\sum_{l \in Y_{L}} Y_{v_{l}} \cdot \ln \left(C \cdot z_{v_{l}}\right)
\end{equation}

where ${Y_{v_l}}$ and ${z_{v_l}}$ are the label and embedding vectors of node ${v_l}$, respectively, and C is the classifier parameter. The overall learning algorithm is outlined in Algorithm \ref{alg1}.

\begin{algorithm}[!h]
	\caption{The overall learning algorithm of LSPI}\label{alg1}
	\begin{algorithmic}[1]
		\Statex \textbf{Input:}the heterogeneous graph ${\mathcal{G}=\{\mathcal{V},\mathcal{E},\mathcal{A},\mathcal{R}}\}$, the initial node feature ${X}$ , region set ${R}$, heterogeneous neighbor set ${\Phi}$, hyperparameter ${\tau}$ and ${T}$, the LSPI model.
		\Statex \textbf{Output:} The node Embeddings $Z$.
		
		\For{Meta-path ${\mathrm{\Phi}_i\in\Phi}$}
		\State Divide the large neighbor path and the small neighbor path by Eq.(\ref{eq1}-\ref{eq4})
		\EndFor
		
		\For {${\Phi_b\in bigPaths}$}	
		\State Calculate the topological probability of the nodes in ${\mathrm{\Phi}_b}$ by Eq.(\ref{eq6})
		\State Calculated feature similarity by Eq.(\ref{eq7},\ref{eq8})
		\State Select the neighbor node in ${\Phi_i}$ by Eq.(\ref{eq9})
		\State Reconstructs the connection relationship in the meta-path by Eq.(\ref{eq10})
		\EndFor
		\State Feature transformation is performed by Eq.(\ref{eq11})
		\For {${\Phi_b\in bigPaths}$}
		\State Compute node embeddings in relation matrix by Eq.(\ref{eq12})
		\EndFor
		
		\For{$\Phi_m\in SmallPaths$}
		\State Perform subgraph convolution operation by Eq.(\ref{eq13})
		\EndFor
		\State Subgraph aggregation level attention, calculate the final embedding ${Z}$ by Eq.(\ref{eq14}-\ref{eq16})
		\State Backpropagation and update parameters according to Eq.(\ref{eq17})
	
	\end{algorithmic}
\end{algorithm}

\section{Experiment}
\subsection{Datasets and Baselines}
This paper selects three widely used heterogeneous graph datasets, ACM, IMDB, and Yelp, to evaluate LSPI. The ACM dataset contains a large number of academic papers covering a broad range of disciplines, with its main nodes consisting of papers (P), authors (A), and subjects (S). The IMDB dataset focuses on information from the movie and television industry, with its main nodes consisting of movies (M), directors (D), and actors (A). The Yelp dataset, a large-scale social media data source widely used in natural language processing and recommendation systems, includes user reviews and ratings of businesses, with its main nodes consisting of businesses (B), users (U), services (S), and rating levels (L). The average degree values and relative difference percentages of the meta-paths in the three datasets are shown in Table \ref{tab1}.

\begin{table}[!h]
	\centering
	\footnotesize
	\caption{Datasets information} 
	\begin{tabular}{|c|c|c|c|c|c|}
		\hline
		Datasets & \multicolumn{1}{c|}{Number of nodes} & \multicolumn{1}{c|}{Linkage} & Meta-path & \makecell[c]{Meta-path average\\ degree value} & \multicolumn{1}{c|}{${R_\Phi}$} \bigstrut\\
		\hline
		\multirow{5}[9]{*}{ACM} & \multirow{5}[9]{*}{\makecell{P:4019\\ A:7167\\ S:60}} & \multirow{5}[9]{*}{\makecell{P-A\\ P-S}} & PAP   & 14.39(min) & 0 \bigstrut\\
		\cline{4-6}          &  &       & PSP   & 1079.42 & 7401.181 \bigstrut\\
		\cline{4-6}          &  &       & PAPAP & 1079.42 & 7401.181 \bigstrut\\
		\cline{4-6}          &       &       & PSPSP & 752.33 & 5128.145 \bigstrut\\
		\cline{4-6}          &       &       & Others & -     & - \bigstrut[t]\\
		\hline
		\multirow{5}[8]{*}{IMDB} & \multirow{5}[9]{*}{\makecell{M:4278 \\ D:2081 \\ A:5257}} & \multirow{5}[9]{*}{\makecell{M-A \\ M-D}} & MAM   & 19.95 & 390.172 \bigstrut[t]\\
		\cline{4-6}          &   &   & MDM   & 4.07(min) & 0 \bigstrut\\
		\cline{4-6}          &   &       & MAMAM & 280.2 & 6784.521 \bigstrut\\
		\cline{4-6}          &       &       & MDMDM & 4.07  & 0 \bigstrut\\
		\cline{4-6}          &       &       & Others & -     & - \bigstrut[t]\\
		\hline
		\multirow{7}[12]{*}{Yelp} & \multirow{5}[9]{*}{\makecell{B:2614 \\ U:1286 \\ S:4 \\ L:9}} & \multirow{5}[9]{*}{\makecell{B-U \\ B-S \\ B-L}} & BUB   & 202.11(min) & 0 \bigstrut[t]\\
		\cline{4-6}          &   &   & BSB   & 947.86 & 368.9822 \bigstrut\\
		\cline{4-6}          &   &   & BLB   & 568.97 & 181.515 \bigstrut\\
		\cline{4-6}          &   &       & BUBUB & 1885.81 & 833.0612 \bigstrut\\
		\cline{4-6}          &       &       & BSBSB & 947.86 & 368.9822 \bigstrut\\
		\cline{4-6}          &       &       & BLBLB & 568.97 & 181.515 \bigstrut\\
		\cline{4-6}          &       &       & Others & -     & - \bigstrut[t]\\
		\hline
	\end{tabular}%
	\label{tab1}%
\end{table}%

In this paper, HAN (2019)\cite{bib10}, MAGNN(2020)\cite{bib11}, HGSL(2021)\cite{bib25}, HPN (2023)\cite{bib12}, ie-HGCN(2023)\cite{bib26}, and SR-HGNN(2023)\cite{bib27}  were selected as the baseline. The performance difference between LSPI and the selected baseline method was compared under different experiments.

\subsection{Experimental Parameter Setup}

This model sets the experimental parameters as follows: middle layer dimension ${d=64}$, learning rate ${lr=0.005}$, optimizer choice Adam, weight decay is ${6.0\times{10}^{-4}}$, feat drop id 0.5, maximum number of iterations is 1000, training and validation set is set to be the total dataset of 10\% and 80\% of the nodes are used for testing.
In terms of meta-paths selection this paper takes into account both meta-paths with different number of neighbors, specifically the ACM dataset selects {PAP, PSPSP, PAPAP} (PSP is the same as PSPSP and therefore only one of them is selected); the IMDB dataset selects {MAM, MAMAM, MDMDM} (MDM is the same as MDMDM and therefore only one of them is selected); the Yelp dataset set chooses {BUB, BUBUB, BSBSB, BLBLB} (BSB is the same as BSBSB and BLB is the same as BLBLB, so only one of them is chosen, respectively).

In terms of large and small neighbor path division, due to the characteristics of different datasets, different parameters ${\tau}$ are set for different datasets. Specifically, for the ACM dataset, with ${\tau=30}$, the large neighborhood paths identified by the path discriminator are {PSPSP, PAPAP}. For the IMDB dataset, with ${\tau=200}$, the large neighborhood path identified is {MAMAM}. For the Yelp dataset, with ${\tau=100}$, the large neighborhood paths identified are {BUBUB, BSBSB}. The large and small neighbor paths corresponding to different ${\tau}$ values are shown in Table II. The number of nodes selected from large neighborhood paths, T, is uniformly set to 500 for all datasets. In part 5.7, this article will study the impact of different ${\tau}$ and T values on the model. In part 5.8, this article will look at not setting the ${\tau}$ value and dividing all paths into large neighbor paths and all into small neighbor paths. Regarding the changes in model performance, this article will further study the selection of T value in Section 5.11. The big and small neighbor paths corresponding to different ${\tau}$ are shown in Table \ref{tab2}.

\begin{table}[!h]
	\centering
	\caption{Neighborhood path segmentation for different ${\tau}$ values of corresponding sizes} 
	\begin{tabular}{|c|c|c|c|}
		\hline
		Datasets & ${\tau}$ & LargePaths & SmallPaths \bigstrut\\
		\hline
		\multirow{2}[4]{*}{ACM} & 30    & PAPAP, PSPSP & PAP \bigstrut\\
		\cline{2-4}          & 100   & PSPSP & PAP, PAPAP \bigstrut\\
		\hline
		\multirow{2}[4]{*}{IMDB} & 100   & MAM, MAMAM & MDMDM \bigstrut\\
		\cline{2-4}          & 200   & MAMAM & MAM, MDMDM \bigstrut\\
		\hline
		\multirow{2}[4]{*}{Yelp} & 70    & BUBUB, BSBSB, BLBLB & BUB \bigstrut\\
		\cline{2-4}          & 100   & BUBUB, BSBSB & BUB, BLBLB \bigstrut\\
		\hline
	\end{tabular}%
	\label{tab2}%
\end{table}%

\subsection{Classification Experiment}
To evaluate the performance of the LSPI model in multi-label classification tasks, the target node features generated by each model were embedded into SVM classifiers with different training ratios and evaluated using Micro-F1 and Macro-F1. The experimental results are shown in Table \ref{tab3}.

\begin{table}[!h]
	\centering
	\scriptsize
	\caption{Classification experiment results}
	\begin{tabular}{|c|c|c|c|c|c|c|c|c|c|c|}
		\hline
		Datasets & Metrics & Split & HAN   & MAGNN & HGSL  & RoHe  & ie-HGCN & HPN   & SR-HGNN & LSPI \bigstrut\\
		\hline
		\multirow{8}[16]{*}{ACM} & \multirow{4}[8]{*}{Macro-F1} & 0.8   & 92.98 & 92.67 & 92.84 & 64.08 & 92.79 & 91.27 & 92.95 & \textbf{94.17} \bigstrut\\
		\cline{3-11}          &       & 0.6   & 92.91 & 92.18 & 92.75 & 93.38 & 92.59 & 91.24 & 92.98 & \textbf{93.89} \bigstrut\\
		\cline{3-11}          &       & 0.4   & 92.72 & 91.39 & 92.59 & 93.86 & 92.14 & 91.08 & 92.71 & \textbf{93.74} \bigstrut\\
		\cline{3-11}          &       & 0.2   & 92.44 & 90.02 & 92.41 & 92.85 & 91.35 & 90.95 & 92.33 & \textbf{93.59} \bigstrut\\
		\cline{2-11}          & \multirow{4}[8]{*}{Micro-F1} & 0.8   & 92.9  & 92.61 & 92.75 & 91.91 & 92.73 & 91.21 & 92.85 & \textbf{94.05} \bigstrut\\
		\cline{3-11}          &       & 0.6   & 92.86 & 92.13 & 92.67 & 93.37 & 92.53 & 91.14 & 92.87 & \textbf{93.77} \bigstrut\\
		\cline{3-11}          &       & 0.4   & 92.67 & 91.38 & 92.53 & 93.32 & 92.11 & 90.98 & 92.62 & \textbf{93.62} \bigstrut\\
		\cline{3-11}          &       & 0.2   & 92.36 & 89.94 & 92.36 & 93.59 & 91.27 & 90.85 & 92.23 & \textbf{93.48} \bigstrut\\
		\hline
		\multirow{8}[16]{*}{IMDB} & \multirow{4}[8]{*}{Macro-F1} & 0.8   & 59.34 & 59.94 & 58.77 & 52.66 & 59.87 & 58.15 & 60.04 & \textbf{63.2} \bigstrut\\
		\cline{3-11}          &       & 0.6   & 59.93 & 59.72 & 58.21 & 55.51 & 59.65 & 58.48 & 59.89 & \textbf{62.22} \bigstrut\\
		\cline{3-11}          &       & 0.4   & 59.7  & 59.23 & 58.02 & 54.98 & 59.33 & 58.42 & 59.54 & \textbf{62.23} \bigstrut\\
		\cline{3-11}          &       & 0.2   & 59.65 & 57.87 & 58.15 & 55.47 & 58.24 & 58.13 & 58.91 & \textbf{61.1} \bigstrut\\
		\cline{2-11}          & \multirow{4}[8]{*}{Micro-F1} & 0.8   & 59.54 & 60.06 & 59.11 & 55.25 & 59.82 & 58.38 & 60.19 & \textbf{63.51} \bigstrut\\
		\cline{3-11}          &       & 0.6   & 60.12 & 59.8  & 58.54 & 56.04 & 59.57 & 58.64 & 60    & \textbf{62.46} \bigstrut\\
		\cline{3-11}          &       & 0.4   & 59.86 & 59.29 & 58.48 & 55.64 & 59.26 & 58.61 & 59.67 & \textbf{62.48} \bigstrut\\
		\cline{3-11}          &       & 0.2   & 59.79 & 57.89 & 58.52 & 55.93 & 58.16 & 58.3  & 59.03 & \textbf{61.39} \bigstrut\\
		\hline
		\multirow{8}[16]{*}{Yelp} & \multirow{4}[8]{*}{Macro-F1} & 0.8   & 71.84 & 92.8  & 93.43 & 93.92 & 91.84 & 90.65 & 90.06 & \textbf{94.49} \bigstrut\\
		\cline{3-11}          &       & 0.6   & 70.27 & 92.64 & 94.41 & 93.13 & 91.83 & 90.41 & 89.88 & \textbf{94.08} \bigstrut\\
		\cline{3-11}          &       & 0.4   & 67.82 & 91.91 & 93.31 & 92.37 & 91.34 & 89.7  & 89.35 & \textbf{93.75} \bigstrut\\
		\cline{3-11}          &       & 0.2   & 63.45 & 90.98 & 93.05 & 92.13 & 91.2  & 88.9  & 88.83 & \textbf{93.15} \bigstrut\\
		\cline{2-11}          & \multirow{4}[8]{*}{Micro-F1} & 0.8   & 81.73 & 91.73 & 92.75 & 94.53 & 90.97 & 90.43 & 90.2  & \textbf{93.87} \bigstrut\\
		\cline{3-11}          &       & 0.6   & 81.02 & 91.49 & 92.62 & 92.88 & 90.85 & 90.09 & 89.93 & \textbf{93.46} \bigstrut\\
		\cline{3-11}          &       & 0.4   & 80.23 & 90.72 & 92.5  & 92.55 & 90.49 & 89.42 & 89.51 & \textbf{93.01} \bigstrut\\
		\cline{3-11}          &       & 0.2   & 78.98 & 89.75 & 92.22 & 91.68 & 90.35 & 88.93 & 89.18 & \textbf{92.43} \bigstrut\\
		\hline
	\end{tabular}%
	\label{tab3}%
\end{table}%

As shown in Table \ref{tab3}, LSPI consistently outperforms baseline methods under different training ratios and achieves significant performance improvements. Specifically, at a training ratio of 80\%, LSPI improves by 1.19\% and 1.15\% on the ACM dataset compared to HAN, by 3.86\% and 3.97\% on the IMDB dataset, and by 22.65\% and 12.14\% on the Yelp dataset. Compared to ie-HGCN, LSPI improves by 3.33\% and 3.69\% on the IMDB dataset and by 2.65\% and 2.9\% on the Yelp dataset. Compared to HPN, LSPI improves by 5.05\% and 5.13\% on the IMDB dataset and by 3.84\% and 3.44\% on the Yelp dataset. We attribute this to LSPI's ability to effectively filter out noisy information before feature aggregation, resulting in higher aggregation quality.

\subsection{Clustering Experiment}

To divide the nodes in the dataset into different clusters or groups, we evaluate the performance of the LSPI model through a node clustering task. The learned node embeddings are used as input to the clustering model. We use the K-means algorithm for clustering and evaluate the clustering performance based on Normalized Mutual Information (NMI) and Adjusted Rand Index (ARI). The experimental results are shown in Table \ref{tab4}.

\begin{table}[!h]
	\centering
	\footnotesize
	\caption{Clustering experiment results}
	\begin{tabular}{|c|c|c|c|c|c|c|c|}
		\hline
		Datasets & Metrics & HAN   & MAGNN & RoHe  & ie-HGCN & SR-HGNN & LSPI \bigstrut\\
		\hline
		\multirow{2}[4]{*}{ACM} & NMI   & 0.6995 & 0.7016 & 0.6756 & 0.4947 & 0.6952 & \textbf{0.7541} \bigstrut\\
		\cline{2-8}          & ARI   & 0.7401 & 0.7214 & 0.6979 & 0.3489 & 0.7465 & \textbf{0.7883} \bigstrut\\
		\hline
		\multirow{2}[4]{*}{IMDB} & NMI   & 0.1201 & 0.1308 & 0.1131 & 0.1308 & 0.1371 & \textbf{0.1471} \bigstrut\\
		\cline{2-8}          & ARI   & 0.1017 & 0.1276 & 0.1079 & 0.1304 & 0.1511 & \textbf{0.1612} \bigstrut\\
		\hline
		\multirow{2}[4]{*}{Yelp} & NMI   & 0.3986 & 0.4734 & 0.5335 & 0.1785 & 0.4998 & \textbf{0.6791} \bigstrut\\
		\cline{2-8}          & ARI   & 0.4461 & 0.3823 & 0.4754 & 0.0639 & 0.5881 & \textbf{0.6994} \bigstrut\\
		\hline
	\end{tabular}%
	\label{tab4}%
\end{table}%

It can be seen from Table \ref{tab4} that LSPI has achieved optimal results in clustering performance on different data sets, especially compared with ie-HGCN, it has achieved a 44\% performance improvement on ACM's ARI and a 26\% performance improvement on NMI. The performance improvement reached 50.06\% and 63.55\% respectively on Yelp; compared with other models, it also achieved significant improvements on the three data sets, which further demonstrates the superiority of the LSPI method.

\subsection{Visualization}

\begin{figure}[!h]
	\centering
	\includegraphics[width=\textwidth]{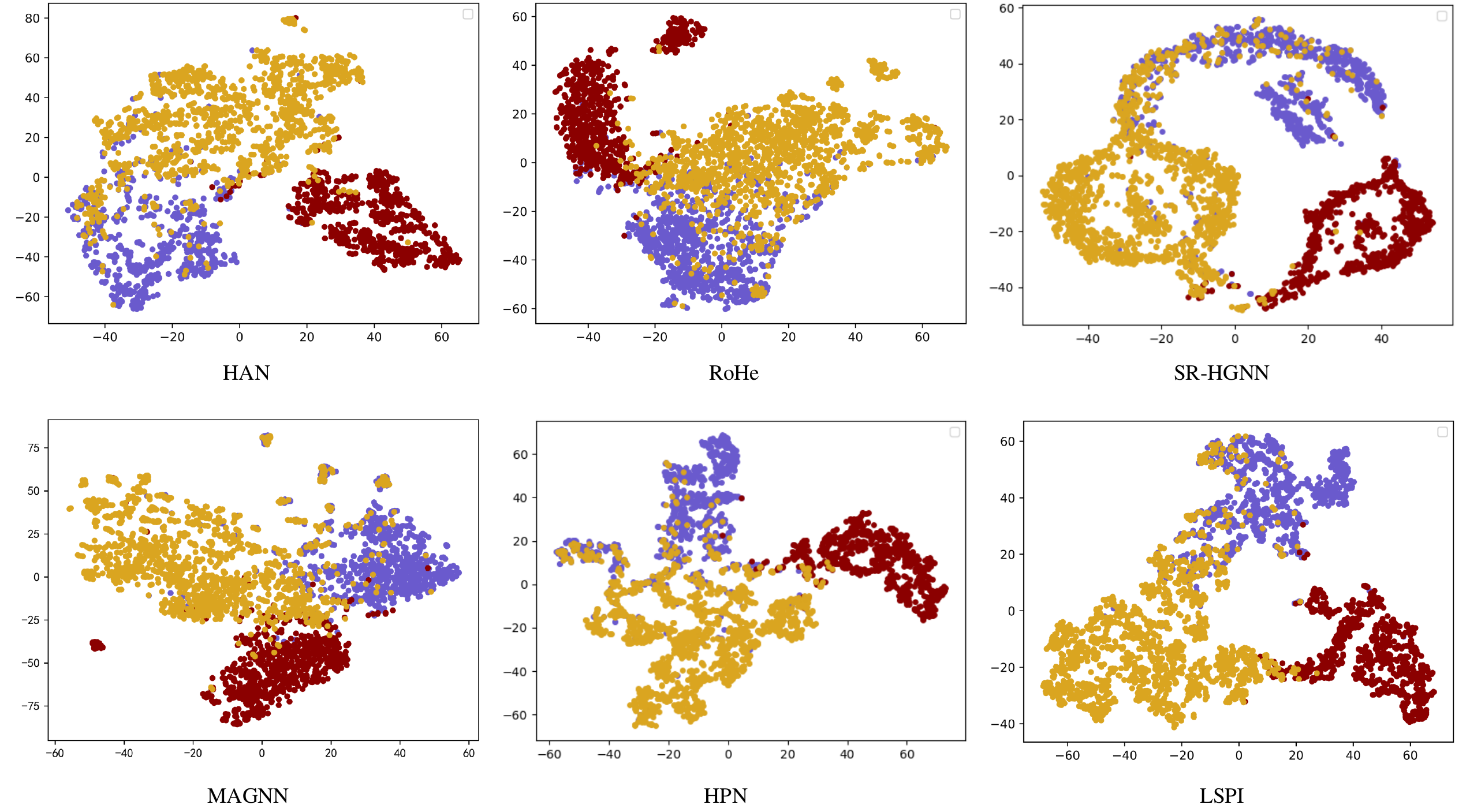}
	\caption{Visualization experiments for node embedding on ACM datasets.}
	\label{fig5}
\end{figure}

In this section, t-distributed stochastic neighbor embedding (t-SNE) \cite{bib28} will be used to map the node embedding of the ACM data set into a two-dimensional space, and three colors will be used to mark different nodes. The experimental results are shown in Figure\ref{fig5}.

\subsection{Ablation Study}

In order to view the performance of different modules of LSPI, this paper deletes the large neighbor path aggregation module (LSPI-w/o-L) and the small neighbor path aggregation module (LSPI-w/o-S) to see the impact of different modules on model performance. After removing one of the modules, all meta-paths will be directly sent to another module for feature aggregation without going through the discriminator. For other settings, refer to Section 5.2.

\begin{figure}[!h]
	\centering
	\includegraphics[width=\textwidth]{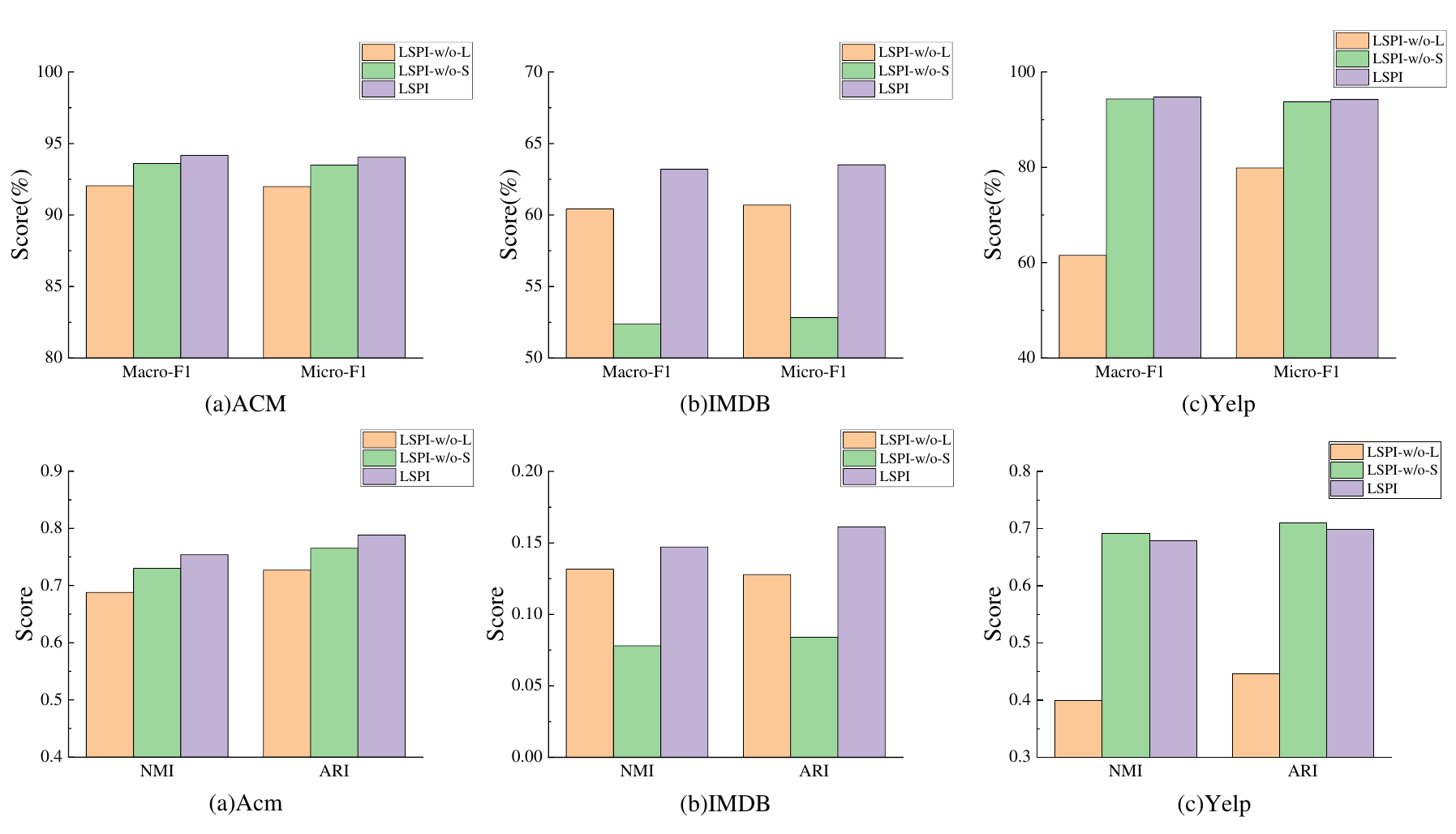}
	\caption{Classification and clustering results under ablation study.}
	\label{fig6}
\end{figure}

As can be seen from Figure \ref{fig6}, the performance of LSPI-w/o-S after de-noising with the large neighbor path module in ACM and Yelp is significantly better than that of LSPI-w/o-L which directly aggregate neighbors based on meta paths, indicating that the large neighbor path aggregation module has achieved remarkable results in removing noise nodes. However, in terms of IMDB data set, the performance of LSPI-w/o-S decreases significantly compared with that of LSPI-w/o-L, indicating that LSPI-w/o-L module also plays a key role in improving model performance. At the same time, the accuracy of LSPI-W /o-B and LSPI-W /o-S on all data sets is lower than that of LSPI, which further proves that the large neighbor path module can effectively shield the noise information in the meta-path, but it is not conducive to preserving the original topological relationship of the HIN. Therefore, it is the best choice to keep both the large neighbor path module and the small neighbor path aggregation module.

\subsection{Parameter Sensitivity Analysis}

\begin{figure}[!h]
	\centering
	\includegraphics[width=\textwidth]{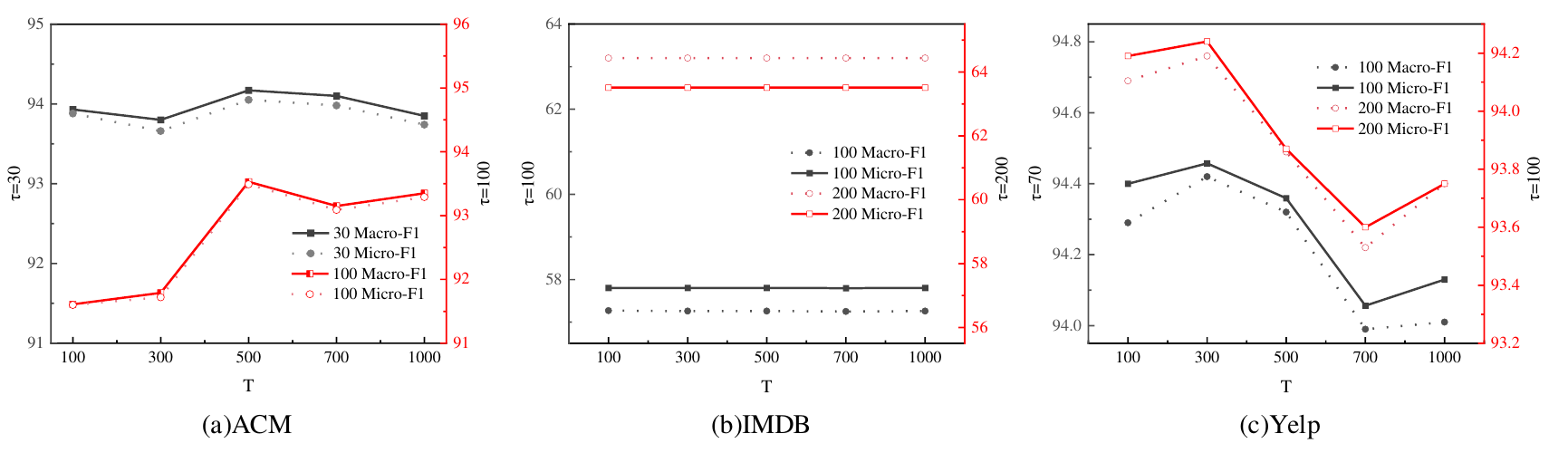}
	\caption{Changes in model accuracy under different $\tau$ values and varying numbers of neighbors $T$.}
	\label{fig7}
\end{figure}

Furthermore, this paper examines the impact of two hyperparameters on the model's performance: the parameter ${\tau}$ for the division of large and small neighborhood paths, and the number of neighbor nodes ${T}$ selected from the large neighborhood paths. ${\tau}$ determines the division between large and small neighborhood paths, while ${T}$ determines the final number of nodes retained in the large neighborhood paths. This paper investigates the effect of different values for these two parameters on model performance. For ${\tau}$, various values are selected to divide the meta-paths into different categories, and the corresponding large and small neighborhood paths for different values are shown in Figure \ref{fig7}. For ${T}$, the values ${\left\{100,300,500,700,1000\right\}}$ are set to observe the impact of different node counts on model accuracy. The experimental results are shown in Fig.6. In Section 5.11, this paper will further study the optimal value for ${T}$ and provide recommendations.

It can be seen from Fig.6 that the division of the path (${\tau}$) and the selection of the number of neighbor nodes (${T}$) have a great impact on the model performance. On the three data sets, the performance is best when the value of ${T}$ is distributed in the 300-500 range. This is because having too few neighbors reduces the model's learning capacity, while having too many neighbors decreases performance due to the increase in noisy nodes. Therefore, the appropriate number of neighbor nodes has an important impact on the performance of the model.
At the same time, it can be seen that the difference in the highest accuracy of the models on the ACM and Yelp data sets is not obvious under different ${\tau}$ values. We believes that this is because the nodes under the LargePaths of the two data sets have strong correlation and less noise information, so the difference in the highest accuracy of the model under different ${\tau}$ values is small. In addition, the IMDB data set performance is similar under different ${T}$ values. This is due to the small number of neighbors in the meta-path and the similar neighbor nodes retained under different ${T}$ values after feature and topology selection.

\subsection{Big Neighborhood Path Performance Test}
Furthermore, this paper examines the performance variations of the model under large neighborhood paths and small neighborhood paths. Specifically, the model variant LSPI-w/o-S is used as the experimental subject. Due to the inability to uniformly measure meta-paths with significant semantic differences, only meta-paths with similar semantics are selected as inputs to observe the performance variations of the model under large neighborhood paths. The selected meta-paths for the three datasets are {PAP, PAPAP}, {MAM, MAMAM}, and {BUB, BUBUB}, respectively. The detailed experimental results are shown in Table \ref{tab5}.

\begin{table}[!h]
	\centering
	\caption{Performance changes under large neighbor paths.}
	\begin{tabular}{|cc|c|c|c|c|}
		\hline
		\multicolumn{1}{|c|}{\multirow{2}[3]{*}{Dataset}} & \multirow{2}[3]{*}{Meta-path} & \multicolumn{2}{c|}{HAN} & \multicolumn{2}{c|}{LSPI-w/o-S} \bigstrut\\
		\cline{3-6}    \multicolumn{1}{|c|}{} &       & Macro-F1 & Micro-F1 & Macro-F1 & Micro-F1 \bigstrut[t]\\
		\cline{2-6}
		\multicolumn{1}{|c|}{\multirow{2}[3]{*}{ACM}} & PAP   & 92.29 & 92.24 & 91.06 & 91.09 \bigstrut[t]\\
		\cline{2-6}    \multicolumn{1}{|c|}{} & PAPAP & 90.24 & 90.29 & 91.98 & 92.02 \bigstrut[t]\\
		\hline
		\multicolumn{2}{|c|}{\textbf{Discrepancy}} & \textbf{-2.05} & \textbf{-1.95} & \textbf{0.92} & \textbf{0.93} \bigstrut\\
		\hline
		\multicolumn{1}{|c|}{\multirow{2}[4]{*}{IMDB}} & MAM   & 52.04 & 53.16 & 52.46 & 52.89 \bigstrut\\
		\cline{2-6}    \multicolumn{1}{|c|}{} & MAMAM & 49.4  & 50.79 & 52.14 & 53.02 \bigstrut\\
		\hline
		\multicolumn{2}{|c|}{\textbf{Discrepancy}} & \textbf{-2.64} & \textbf{-2.37} & \textbf{-0.32} & \textbf{0.13} \bigstrut\\
		\hline
		\multicolumn{1}{|c|}{\multirow{2}[4]{*}{Yelp}} & BUB   & 63.95 & 72.02 & 92.42 & 91.51 \bigstrut\\
		\cline{2-6}    \multicolumn{1}{|c|}{} & BUBUB & 77.83 & 74.86 & 92.44 & 91.59 \bigstrut\\
		\hline
		\multicolumn{2}{|c|}{\textbf{Discrepancy}} & \textbf{13.88} & \textbf{2.84} & \textbf{0.02} & \textbf{0.08} \bigstrut\\
		\hline
	\end{tabular}%
	\label{tab5}%
\end{table}%

\subsection{Robustness Study}

\begin{figure}[!h]
	\centering
	\includegraphics[width=\textwidth]{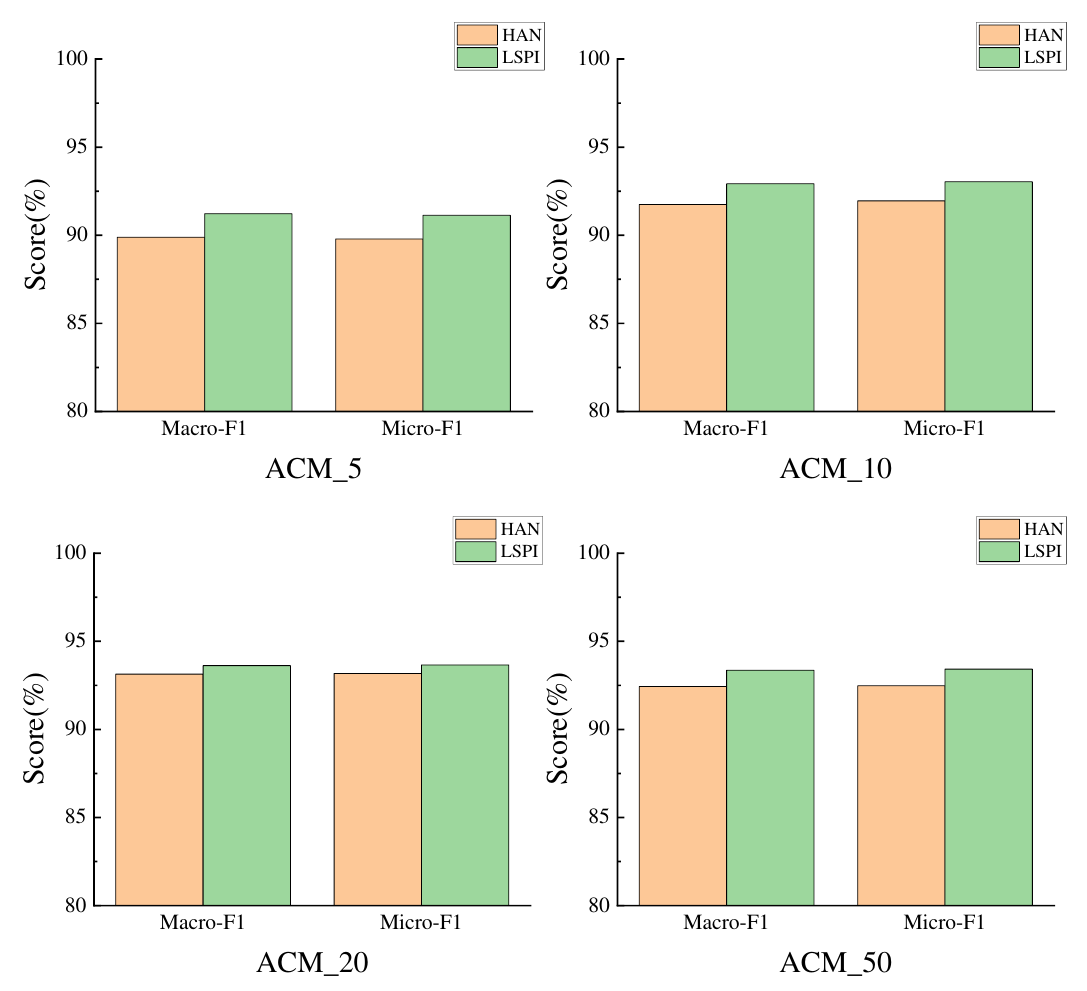}
	\caption{Experimental results of LSPI and HAN on randomly deleting part of the node data set.}
	\label{fig8}
\end{figure}

Considering the removal of neighbor nodes in the large neighborhood path module, this section further verifies the robustness of the model. We randomly deleted ${1 / 5,1 / 10,1 / 20,1 / 50}$ of the nodes and their adjacent edges in the ACM dataset to create four new datasets, denoted as ACM\_5, ACM\_10, ACM\_20, and ACM\_50. These datasets were then input into LSPI and HAN to test the robustness of the models The detailed experimental results are shown in Figure \ref{fig8}.

From Figure \ref{fig8}, it can be observed that the more nodes are deleted, the more significant the performance decline of the model. However, across all four experimental results, LSPI consistently outperforms HAN. Additionally, it can be seen that the rate of decline for LSPI is significantly lower than that of HAN, indicating that LSPI has stronger resistance to interference.

\subsection{Node Number Research}
To provide the optimal reference values for T under different datasets, this paper, based on Section 5.7, selects the highest accuracy value of ${\tau}$, and further studies the model performance using four criteria(D\_Max, D\_Min, D\_Avg, D\_Med). These criteria represent the maximum degree, minimum degree, average degree, and median degree of all large neighborhood paths selected under the designated ${\tau}$ value. The specific information is shown in Table\ref{tab6}. In the experiments, the decimal places of D\_Avg will be rounded.

\begin{table}[!h]
	\centering
	\caption{The large neighborhood paths selected and their corresponding indicator values under different datasets.}
	\begin{tabular}{|c|c|c|c|c|c|c|}
		\hline
		Datasets & ${\tau}$     & Big\_Path & D\_Max & D\_Min & D\_Avg & D\_Med \bigstrut\\
		\hline
		\multirow{2}[4]{*}{ACM} & \multirow{2}[4]{*}{30} & PAPAP & \multirow{2}[4]{*}{2595} & \multirow{2}[4]{*}{2} & \multirow{2}[4]{*}{1172.82} & \multirow{2}[4]{*}{900} \bigstrut\\
		\cline{3-3}          &       & PSPSP &       &       &       &  \bigstrut\\
		\hline
		IMDB  & 200   & MAMAM & 1365  & 1     & 280.2 & 190 \bigstrut\\
		\hline
		\multirow{2}[4]{*}{Yelp} & \multirow{2}[4]{*}{100} & BUBUB & \multirow{2}[4]{*}{3692} & \multirow{2}[4]{*}{171} & \multirow{2}[4]{*}{2833.68} & \multirow{2}[4]{*}{2952} \bigstrut\\
		\cline{3-3}          &       & BSBSB &       &       &       &  \bigstrut\\
		\hline
	\end{tabular}%
	\label{tab6}%
\end{table}%
The experimental results are shown in Table \ref{tab7}. Although the model accuracy under the four criteria did not exceed the highest value in Section 5.7, it can be observed that when ${T}$ is set to D\_Avg and D\_Med, the model achieves higher accuracy scores across all datasets. Therefore, this paper suggests that ${T}$ should lean towards the average degree and median degree of large neighborhood paths.

\begin{table}[!h]
	\centering
	\caption{Experimental results with different indicators.}
	\begin{tabular}{|c|c|c|c|c|c|}
		\hline
		Datasets & Metric & ${T=}$D\_Max & ${T=}$D\_Min & ${T=}$D\_Avg & ${T=}$D\_Med \bigstrut\\
		\hline
		\multirow{2}[3]{*}{ACM} & Macro-F1 & 93.9  & 91.79 & 93.79 & 93.86 \bigstrut[t]\\
		& Micro-F1 & 93.82 & 91.75 & 93.65 & 93.73 \bigstrut[t]\\
		\hline
		\multirow{2}[2]{*}{IMDB} & Macro-F1 & 63.2  & 61.94 & 63.2  & 63.2 \bigstrut[t]\\
		& Micro-F1 & 63.51 & 62.13 & 63.51 & 63.51 \bigstrut[t]\\
		\hline
		\multirow{2}[2]{*}{Yelp} & Macro-F1 & 94.45 & 94.7  & 94.47 & 94.45 \bigstrut[t]\\
		& Micro-F1 & 93.77 & 94.14 & 93.79 & 93.77 \bigstrut[t]\\
		\hline
	\end{tabular}%
	\label{tab7}%
\end{table}%

\section{Conclusion}

This paper addresses the challenge of significant discrepancies in the number of neighbors across different meta-paths and the presence of noise in large neighborhood paths. To tackle these issues, a heterogeneous graph neural network algorithm based on the discrimination of large and small neighborhood paths is proposed, named LSPI. LSPI first divides meta-paths into large and small neighborhood paths using a path discriminator, and then selects nodes from the large neighborhood paths based on both topology and features to mitigate noise interference. Subsequently, feature information from different paths is obtained through a graph convolutional module and fused using a graph-level attention mechanism. Comprehensive experimental results demonstrate that LSPI exhibits favorable model performance and significant improvement in handling large neighborhood paths.

\section*{Acknowledgements}
This work is supported by National Key R\&D Program of China [2022ZD0119501]; NSFC [52374221]; Sci. \& Tech. Development Fund of Shandong Province of China [ZR2022MF288, ZR2023MF097]; the Taishan Scholar Program of Shandong Province[ts20190936].



\bibliographystyle{elsarticle-num} 
\bibliography{ref}


\end{document}